\documentclass{article}
\usepackage{spconf}
\usepackage{amsmath,graphicx}
\usepackage{xcolor}
\usepackage[justification=centering]{subcaption}
\usepackage{lipsum}
\graphicspath{{images/}}

\title{Parcellation of Visual Cortex on high-resolution histological Brain Sections using Convolutional Neural Networks}
\name{Hannah Spitzer$^{1}$ \qquad Katrin Amunts$^{1,2}$ \qquad Stefan Harmeling$^{3}$ \qquad Timo Dickscheid$^{1}$}
\address{
	$^{1}$ Institute of Neuroscience and Medicine (INM-1), Forschungszentrum J\"ulich, Germany\\
	$^{2}$ C. and O. Vogt Institute for Brain Research, Heinrich-Heine University D\"usseldorf, Germany\\
	$^{3}$ Institut f\"ur Informatik, Heinrich-Heine University D\"usseldorf, Germany
}
\begin{document}
%\ninept
%
\maketitle
\begin{abstract}
	% 100 - 150 words
	Microscopic analysis of histological sections is considered the ``gold standard" to verify structural parcellations in the human brain. 
	Its high resolution allows the study of laminar and columnar patterns of cell distributions, which build an important basis for the simulation of cortical areas and networks. 
	However, such cytoarchitectonic mapping is a semi-automatic, time consuming process that does not scale with high throughput imaging. 
	We present an automatic approach for parcellating histological sections at $2\,\mu m$ resolution. 
	It is based on a convolutional neural network that combines topological information from probabilistic atlases with the texture features learned from high-resolution cell-body stained images.
	The model is applied to visual areas and trained on a sparse set of partial annotations. We show how predictions are transferable to new brains and spatially consistent across sections. 
	
	%With the increasing amount of high resolution histological data available through high-throughput scanning, the current semi-automatic workflow for mapping needs to be supported by automatic methods.
	%We present an automatic approach for parcellating histological sections at $2\,\mu m$ resolution combining coarse global anatomical estimates from probabilistic atlases with precise local features from high-resolution cell-stained images. 
	%Our convolutional neural network model is trained on a sparse set of partial annotations of the visual system and is transferable to new brains and spatially consistent across sections.
	%We show the precise contribution of the probabilistic atlases to the segmentation result.

\end{abstract}

\section{Introduction}
\label{sec:intro}
%
% --- What is the problem and why is it interesting ---
Precise delineations of cytoarchitectonic areas in cell-body stained histological sections of the human brain provide a basis for a multimodal brain atlas.
They are indispensable for allocating the multiscale functional imaging, physiological, connectivity, molecular, and/or genetic data to anatomically well specified entities of the human brain organization at high spatial resolution \cite{amunts2015}. 
% --- Why is it hard ---
Cytoarchitectonic areas are distinguished by variations of the cell distribution in the cortical laminae and with respect to the columnar organization of the cortex. 
Parcellation of cortical areas therefore requires an image resolution of $1$-$2\,\mu m$ to distinguish individual neurons, and to capture their morphology. 
Image analysis and multivariate tools have been introduced to detect boundaries of cortical areas in a reproducible semi-automatic way \cite{schleicher1999}.
This method, however, is time- and labor-intensive, and significantly constraints mapping efforts in a large sample of histological sections and/or brains. 
Thus, methods with higher degree of automation are needed.  
%To avoid bottlenecks, they should scale with (high-throughput) imaging of histological sections 

% --- What we do ---
This paper proposes a convolutional neural network (CNN) model that parcellates high-resolution sections from different subjects by exploiting 1) prior knowledge about reasonable topologies as given by existing probabilistic atlases, and 2) precise local features extracted from the cell-stained tissue scan.
Our model is trained on partial delineations and produces parcellations that are transferable to new brains and spatially consistent across sections. 

%Our model is trained on partial delineations and produces validated and spatially consistent parcellations. 
%Our model produces parcellations that are spatially consistent across sections and is applicable to brains that were not in the training set.
% mention that we want to help mapping process?

% --- Related work, and why is out method different than previous methods ---
After the recent success of CNNs for natural image classification, approaches for efficient semantic segmentation with CNNs were developed (e.g., \cite{ronneberger2015}). 
There are first works proposing CNN-based models to parcellate entire 3D MR volumes according to different segmentation protocols \cite{brebisson2015, lee2011}. 
In contrast, we aim towards parcellating high-resolution cell-stained histological sections at the quality of a purely cytoarchitectonic reference parcellation.
%todo "at the quality of"... alternatives for this?
% lee: towards a deep learning approach to brain parcellation
% mention that we only want to do 2d, because 3d is not yet clear how it works on this resolution?

% --- Contributions ---
To our knowledge, our approach is the first to tackle automatic parcellation of cortical areas in histological sections. 
This paper makes the following contributions: 
%(i) 
%mention: show architecture of such model?
We train a model that automatically predicts 13 areas of the human visual system in histological sections.
%(iii)
To deal with the fact that we only have access to partially delineated sections for training,
we automatically create accurate gray/white matter segmentations using the same CNN architecture, 
and use them to distinguish between ``unlabeled cortex" and background.
%These gray/white matter segmentations are themselves an important result of our work. 
%todo too detailed?
%(ii)
We evaluate the influence of exploiting probabilistic atlases on the anatomical correctness of the automatic parcellation, and show results indicating that the model is transferable to previously unseen brains and consistent across multiple consecutive sections in the same brain.

%--Alternative
%Additionally, we show how we deal with partial groundtruth labels to enable parcellation of entire brain sections. 
%This procedure enabled us to generate accurate white matter / gray matter segmentations with very few training data.

%Recently, there have been severeal approaches for using CNNs to do efficient semantic segmentation in natural images \todo{cite caffe fcn, unet, dilated, deconv}. All of these approaches have in common that they seek a way to relate far away pixels with each other and keep the fine-grained spatial dependencies needed for a precise segmentation. These approaches are a good starting pouint for designing a model for parcellating histologic sections. 
%In the realm of human brain segmentation, some recent works also appear: \cite{brebisson} \todo{more}. These approaches are trained for parcellating (high resolution) 3D MR data, whereas we want design a model for parcellating high-resolution post-mortem cellstained histological sections, wihtouhg requireing an computationally expensive registration in a 3D volume first. 

\section{Model}
\label{sec:arch}
\subsection{Network Architecture for Semantic Segmentation}
\label{sec:arch:arch}
\vspace*{-.5\baselineskip}

\begin{figure}[t]
	\includegraphics[width=\columnwidth]{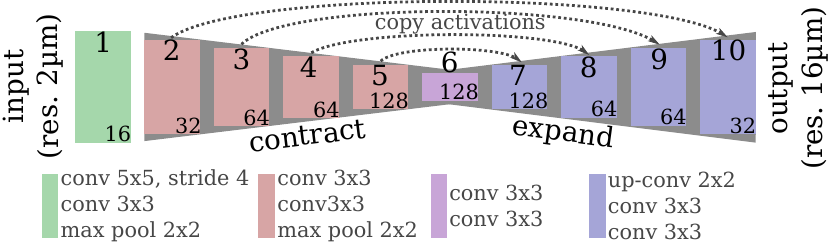}
	\caption{
		\textbf{Base net}, based on \cite{ronneberger2015}, consisting of 10 blocks containing several layers (denoted by colored rectangles). 
		The number of channels of the layers is shown at the bottom of each block. 
		%Blocks summarizing two convolutional and a pooling or upsampling layer are denoted with rectangles. 
		%The input block $1$ applies a $5\times5\times16$ convolution with stride $4$, followed by a $3\times3\times16$ convolution and a $2\times2$ max-pooling layer. 
		%The following blocks contain $3\times3$ convolutions with varying filter sizes (32,64,64,128,128,128,64,64 filters respectively). 
		%
		%Layers 2-5 comprise the compressing branch: two convolutional and a pooling layer each; 32, 64, 64, 128 filters respectively. 
		%Layer 6 is at the bottom of the net: two convolutional layers; 128 filters. 
		%Layers 7-9 comprise the upscaling branch: a deconvolution layer and two convolutional layers each; 128, 64, 64 filters respectively. 
		%Layers 2-5 (compressing branch) contain a max pooling layer, and layers 7-9 (decompressing branch) a deconvolution layer.
		%Convolutions have size $3\times3$ and batch normalization is applied after each convolution.
		Batch normalization and ReLU are applied after each convolution.
		%Activations are copied from the contracting path and appended to the expansive path (dashed lines). 
		Total number of parameters: $1479728$. Receptive field size: $1481^2\,px^2 \approx 3^2\,mm^2$. 
		%Block denotes a compound of convolutional layers and a pooling or deconvolution layer
	}
	\label{fig:arch}
\end{figure}
%
% --- UNet ---
We base our CNN architecture on the semantic segmentation approach of \cite{ronneberger2015}. 
They designed a network with a contracting path, consisting of several ``blocks" that contain several convolutional layers and one pooling layer to downsample the activations each. 
This is followed by an expansive path with ``blocks" of one upsampling layer and several convolutional layers.
The activations from the contracting path are appended to the expansive path to enable the propagation of context information to higher resolution layers. 
%To ensure a precise final segmentation, the activations from the contracting path are appended to the expansive path which upscales the activations and enables the propagation of context information to higher-resolution layers. The "blocks" on the expansive path contain one upsampling layer and several convolutional layers. %using dilated convolutions
This architecture produces precise segmentations for any input size and is thus suited for efficient parcellation of entire brain sections. %additional plus: low memory consumption

% --- Justification for changes in architecture ---
Cytoarchitectonic parcellation of the cortex relies on the size and composition of cortical laminae \cite{schleicher1999}. 
%Traditionally, this is done by comparing statistics of profiles traversing the cortex...
Thus for the classification of any pixel inside the cortex, we choose to take the whole depth of the cortex into account. 
%For CNNs this means that the receptive field should at least encompass the entire cortex. 
Assuming a cortical depth of $2$-$4\,mm$ and an input resolution of $2\,\mu m$, the receptive field of the CNN should be about $1000^2$-$2000^2\,px^2$. %special attention is needed, larger than most/all other networks. 
%For one whole brain section of the visual cortex (average size 60mm x 100mm) this amounts to $6*10^9$ pixels that need to be processed. 
%Thus any model that parcellates cortical areas in high resolution brain sections needs to be both time and memory efficient in order to deal with when predicting and feasible to train. 

% --- Our adaption to base model ---
Following these considerations, we increase the receptive field of the network by inserting one ``block" containing convolutional layers with stride 4 before the contracting path. 
Although a $2\,\mu m$ resolution is needed to see the relevant cytoarchitectonic features for mapping, the expected practical localization accuracy of the resulting cytoarchitectonic borders is much lower.
We take this into account by setting the output resolution of our network to $16\,\mu m$, which has the practical benefit of significantly reducing the memory requirements of the model during training.
%Since the estimated precision of cytoarchitectonic borders is coarser than $2\,\mu m$, parcellation results do not need to be at this resolution. 
%In our architecture the output resolution is $16\,\mu m$, which has the additional benefit of significantly reducing the memory requirements of the model during training. 

%However, if a sufficiently large GPU is available, extending the expansive path by one "block" is trivial.
% --- training hints (but most are later in results) ---
In order to train converging networks we found it necessary to add batch normalization after each convolutional layer and use a relatively high learning rate. 
Fig.~\ref{fig:arch} shows details of our network architecture.  

\vspace*{-.5\baselineskip}
\subsection{Exploiting probabilistic atlas information}
\label{sec:arch:prior}
\vspace*{-.5\baselineskip}

%todo{using net and model interchangeably - stick to one!}
% --- Motivation why use pmaps ---
To identify a certain area in the brain, neuroscientists first identify a region of interest (ROI) on a low-resolution global view and then map the precise borders of this area by means of local texture patterns on a high-resolution local view.
%In particular the global view helps to disambiguate areas that have very similar texture but are located in different locations of the brain. 
In a similar manner, our model combines precise local features from high-resolution cell-stained tissue with a relatively imprecise but topologically correct probabilistic atlas prior. 
We use the JuBrain atlas\footnote{available at http://www.jubrain.fz-juelich.de/} which gives our model probability maps for each area, helping to disambiguate areas that have similar texture but are located in different locations of the brain. 

\begin{figure}
	\includegraphics[width=\columnwidth]{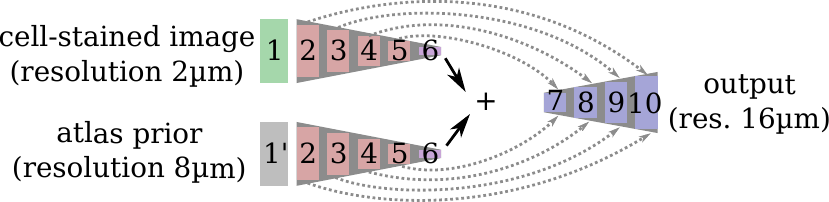}
	\caption{
		\textbf{Atlas-aware net} including transformed probabilistic atlas (atlas prior) in an additional contracting path.
		Activations from both contracting paths are appended to the joint expansive path (dashed lines).
		Input block $1'$ for the atlas prior consists out of one $5\times5\times16$ convolution (no pooling).
	}
	\label{fig:arch-fusion}
\end{figure}

%old
%In the past, a lot of effort has gone towards creating high-resolution probabilistic maps in the colin27 reference space \cite{amunts}. 
%Including these maps\footnote{available at LINK} in our model enables it to access the expert knowledge of how probable it is that certain areas are located in the input image.
%This preempts confusions of cortical areas that are very far from each other, because the model has a rough estimate of where in the brain the input image is located. %in combination with the local texture cues from the cell-stained input image. 

% --- Registration of pmaps ---
Using the probabilistic atlas requires a registration from the atlas space to the space of the individual brain. 
Since we are only interested in rough estimates of the atlas data on the individual section, we calculated an affine registration from the atlas to the individual sections at $8\,\mu m$ resolution.
An example of projected probabilities for one area is depicted in Fig.~\ref{fig:res}a.
%shows the projected probability of area hOc1 overlayed with the groundtruth delineations on one section. 
%
% --- extended architecture ---
%For every area that should be predicted by the model, the projected atlas prior is added as an additional input.
%Theoretically, the pmaps could be included at any layer of the network, as long as the scales of the gray-level activations at this layer and of the pmap coincide. 
We add a second contracting path for the atlas data of every area that should be predicted by the model, and join the resulting activations to the bottom of the cell-stained image expansive path (Fig.~\ref{fig:arch-fusion}). 
This enables the network to process each input type individually in the contracting paths and jointly in the expansive path. 

% --- avoid that only pmaps are predicted ---
We believe that the network learns the atlas prior faster than the image of the cell-body stained section, because the atlas data are less complex and directly represent an estimate of the output labels.
%When adding the pmaps to the inputs of the net we basically give it an estimate of the labels it should predict. 
This bears the danger that the model converges to an inferior solution, which is the local optimum of predicting a linear combination of the probabilistic maps without learning the underlying texture pattern contained in the cell-stained image. 
In particular, we observed this behavior in architectures that were joining the atlas and cell-stained image paths earlier in the contractive paths. %in the network 
To overcome this problem, we add noise to the atlas input (set every input node to $0$ with a chance of $20\%$) and use an iterative training procedure: 
First, the model is trained on only the cell-stained image and only after convergence training is continued using both cell-stained image and atlas information. 
This procedure ensures that the model learns to use the information contained in the cell-stained image first, and then adapts to include information gained from the atlas. 

\vspace*{-.5\baselineskip}
\subsection{Data preparation}
\label{sec:arch:train}
\vspace*{-.5\baselineskip}

%Following the aim of a model that is easy to apply to 2D histological sections without requiring a time-intensive registration step beforehand, we train the model directly on the gray level images. \todo{will not mention bias correction, since this did not have any improvement!} 
% --- laplace rotation ---
The learned model should be robust to the local shape of the cortex (e.g., gyri, sulci) and recognize cortical areas by their texture. 
To support this, we normalize the input data by rotating each input crop along the main direction of the gradient of the Laplacian field between outer and inner cortical boundary (using the segmentation from Sec.~\ref{sec:seg}). 
This ``Laplacian field orientation correction" aligns all input data and makes the model more applicable to new subjects (see Sec.~\ref{sec:res}). In our evaluation, models trained with this rotation of input data performed about two Dice coefficient points (a.k.a. F1 score) better than models trained with random rotation of input data. This mildly improves the results. 

\vspace*{-.5\baselineskip}
\subsection{Dealing with complex background class}
\label{sec:arch:labels}
\vspace*{-.5\baselineskip}

% -- Motivation: fill up training data ---
% --- Description labels ---
As groundtruth labels we use partial delineations of visual areas on high-resolution brain sections. 
We can neither expect all visual areas to be labeled in an annotated section (partial annotation), nor do we have access to groundtruth annotations in all consecutive sections (sparse annotation).
%Not the entire brain section is labeled (partial groundtruth) and not all visual areas are annotated on all sections (missing groundtruth). 
Thus the background class is a compound of very different concepts: white matter (\emph{wm}), non-visual or non-labeled cortex (unknown cortex, \emph{gm}), and background (\emph{bg}). 
This makes it much harder for the net to learn the background class. 
Thus, we split up the background label in its distinct concepts \emph{wm}, \emph{gm}, and \emph{bg} (see Sec.~\ref{sec:seg}), and extend the groundtruth to include the classes ``unknown cortex" and ``white matter".

\section{Gray / White Matter Segmentation}
\label{sec:seg}

\begin{figure} %[htb]
		\centering
		\begin{subfigure}[t]{.38\linewidth}
			\centering
			\includegraphics[width=.95\linewidth, trim=150 0 100 150, clip]{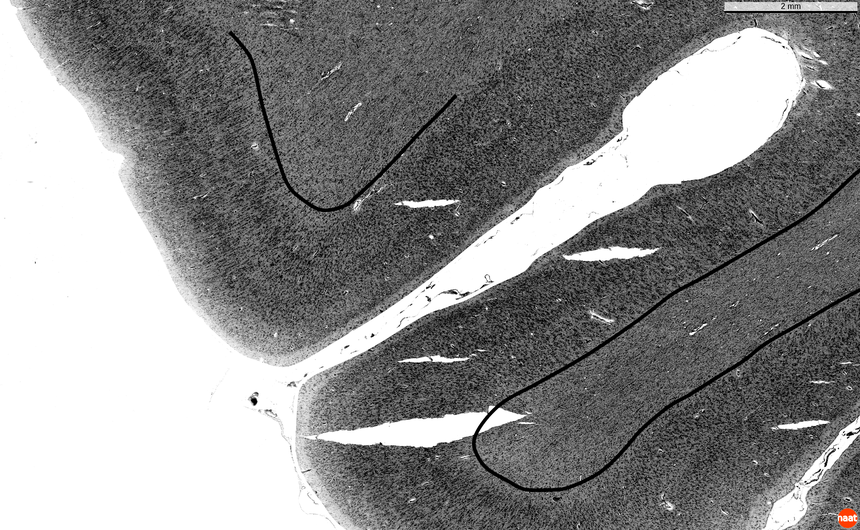}
			%\caption{input image}
			%\label{}
		\end{subfigure}%
		\begin{subfigure}[t]{.38\linewidth}
			\centering
			\includegraphics[width=.95\linewidth, trim=150 0 100 150, clip]{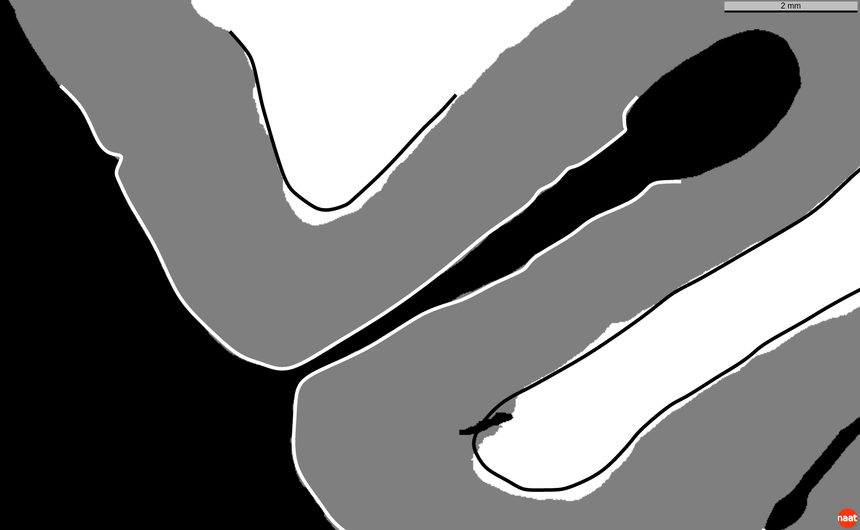}
		%	\caption{segmentation result}
		%	\label{}
		\end{subfigure}
		\caption{
			\textbf{\emph{Gm/wm/bg} segmentation}, with input image left and segmentation result right.
			Solid lines represent manual delineations of inner (black) and outer (white) cortical boundary.
		}
		\label{fig:bgseg}
\end{figure}

% --- Motivation ---
For the Laplacian field orientation correction (Sec.~\ref{sec:arch:train}) and the extension of partial annotations (Sec.~\ref{sec:arch:labels}), a reliable gray/white matter segmentation is needed.
%
% --- Process to get to segmentation ---
We have devised a two-step procedure with minimal labeling overhead to train a CNN model with the base architecture (Sec.~\ref{sec:arch:arch}) for this task.
First, we train a model for the two-class cortex segmentation task. 
As groundtruth all available delineations of the visual cortex were set as ``cortex" and the remaining pixels as ``background". 
Obviously, the generated background class also contains (non-visual) cortex. 
Thus, we adapt the loss function to weigh the error ``predict cortex, true background" with $0.5$ while weighing the other error with 1. 
In the second step, we labeled $20$ sections with \emph{wm}, \emph{gm}, and \emph{bg}, by manually inserting \emph{wm} labels in the cortex segmentations of the previous step and trained a model on these sections. %for segmenting white matter, background and cortex. 

% --- Quality of the result ---
%The resulting segmentation has a resolution of $8\,\mu m$ and is to our knowledge the first automatic segmentation on the high resolution post-mortem histological brain sections. 
The resulting three-class segmentation has a resolution of $8\,\mu m$, is robust to small cuts and rips in the tissue and to noise in the background and sets a precise outer cortical boundary (Fig.~\ref{fig:bgseg}). 
Compared to expert segmentations, the automatic segmentation is consistent and reproducible, but seems to systematically underestimate the inner cortical boundary (especially in the curves). 
%However, it is a good result given the sparse and low-resolution training data (delineations of areas at $20\,\mu m$).

\section{Results}
\label{sec:res}

\begin{figure}[tb]
	\centering
	\begin{subfigure}[t]{.28\columnwidth}
		\centering
		\includegraphics[width=.98\linewidth]{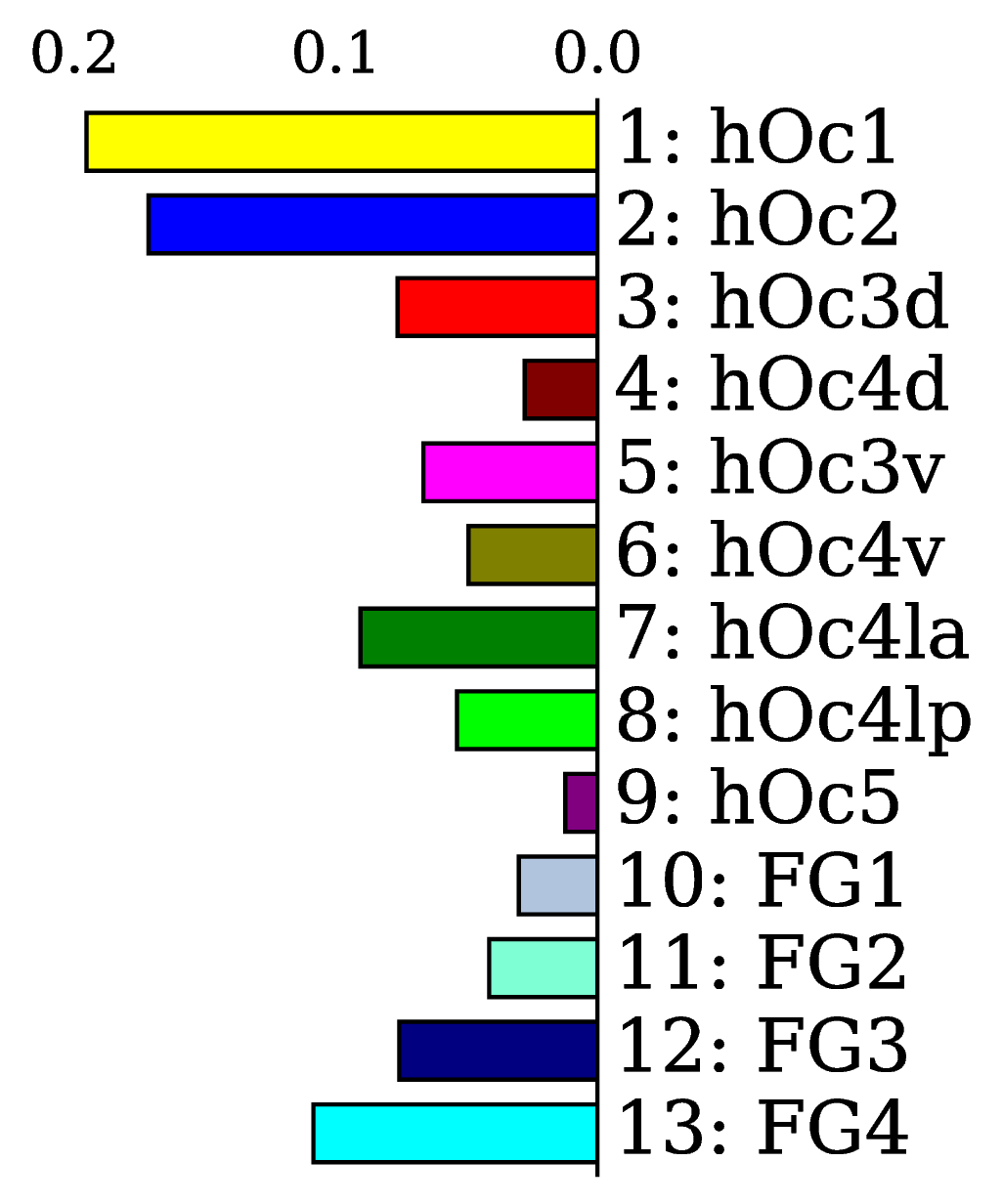}
		\caption{label frequency in dataset}
		\label{fig:hist}
	\end{subfigure}
	\begin{subfigure}[t]{.34\columnwidth}
		\centering
		\includegraphics[width=.98\linewidth]{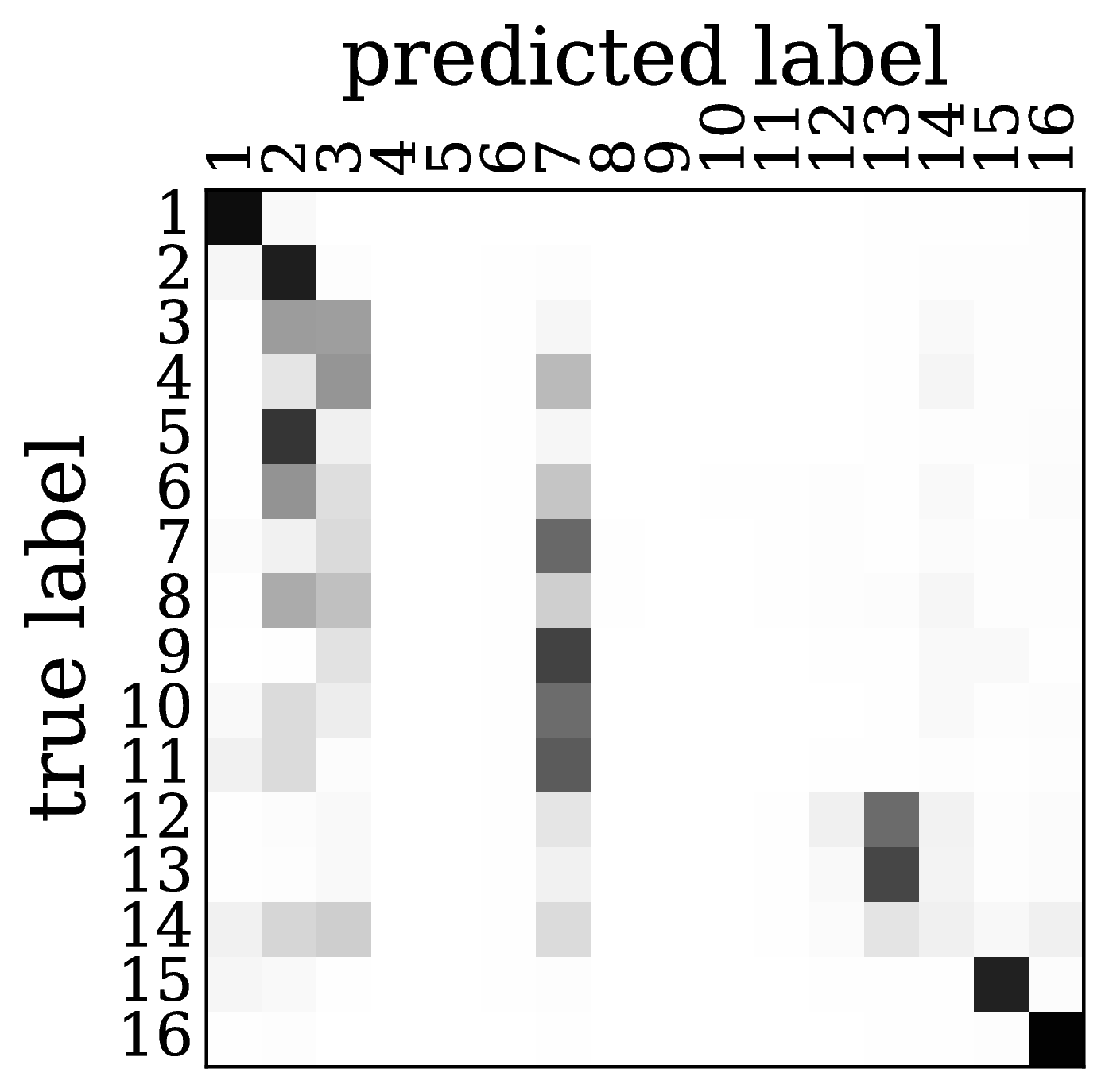}
		\caption{base: $D_C=0.62$, $\epsilon=23.3$}
		\label{}
	\end{subfigure}
	\begin{subfigure}[t]{.36\columnwidth}
		\centering
		\includegraphics[width=.98\linewidth]{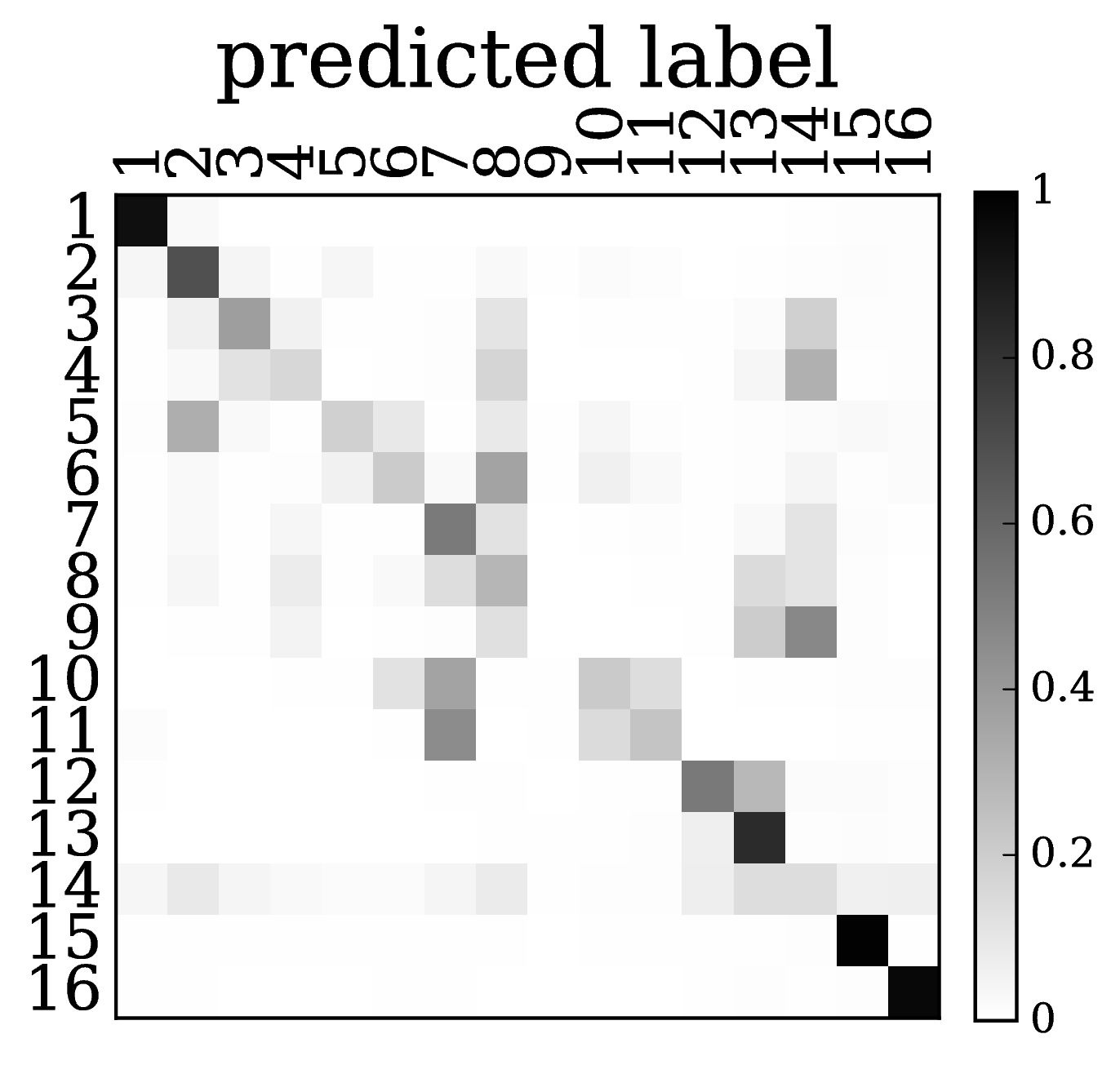}
		\caption{atlas-aware: $D_C=0.72$, $\epsilon=21.2$}
		\label{}
	\end{subfigure}
	\caption{
		\textbf{Quantitative evaluation of base and atlas-aware model}. 
		Label numbers 14-16 denote labels \emph{gm}, \emph{wm}, and \emph{bg}.
		}
	\label{fig:cm}
\end{figure}

%This section describes the quantitative and qualitative evaluations show that our model is anatomically correct, spatially consistent and transferable to other brains. 

\textbf{Dataset.} 
The dataset used for training and validating the models contains in total $111$ cell-body stained histological sections of the human visual cortex originating from four different brains. 
Mapping resulted in $13$ areas (primary and higher visual areas of the dorsal and ventral stream), with an average of $5$ mapped areas per section. 
Fig.~\ref{fig:hist} shows the frequency of the areas in the dataset.
As described in Sec.~\ref{sec:arch:labels}, the groundtruth was extended by automatic segmentations of unknown cortex, white matter, and background. 

\textbf{Training.} 
% --- Samping method ---
The models were trained on $2/3$ of the sections ($1/3$ held out for test and validation) by drawing patches of size $2000\times2000$ from the images ($85\,\%$ from the delineated areas, $15\,\%$ from background). 
% --- Training details ---
We used a learning rate of $0.05$ and a batch size of $20$. Usually, training converged after $5000$ iterations.
% --- Remark on overfitting (response to review) ---
Although the number of training sections is small, we do not observe any overfitting. We attribute this to the random sampling during training which never produces the same patch twice, and the difficulty of the task. The trained model has expressive filters in the first layers.

\textbf{Evaluation.} 
Quantitative evaluation scores were computed on the held-out test set. 
The standard score to assess the quality of a segmentation is the mean Dice coefficient $D_C$ (e.g., \cite{lee2011, brebisson2015}).
However, this does not take into account the spatial information of the segmentation. 
For anatomical parcellation, an error made right at the border between two areas is ``less severe" than confusing two areas that lie in different regions of the brain. 
Thus we additionally report the pixel distance error ($\epsilon$) which assigns to each error a penalty based on the distance between the misclassified pixel and the nearest groundtruth pixel that actually is of the misclassified class \cite{yasnoff1977}:
$\epsilon_{\tau} = \sum_{i=1}^{N} d^2(i)$, with $d^2(i)$ the squared Euclidean distance of the $i$th misclassified pixel to the nearest true pixel with this class. 
Following \cite{yasnoff1977}, the error is normalized to range $0$ and approx.\;$100$, so it can be understood as a percentage of the maximum possible error: 
$\epsilon = 100*\frac{\sqrt{\epsilon_{\tau}}}{A}$, with $A$ being the total number of pixels that were evaluated.

\begin{figure*}[t!]
	\centering
	\begin{subfigure}[t]{.19\linewidth}
		\centering
		\includegraphics[width=.97\linewidth]{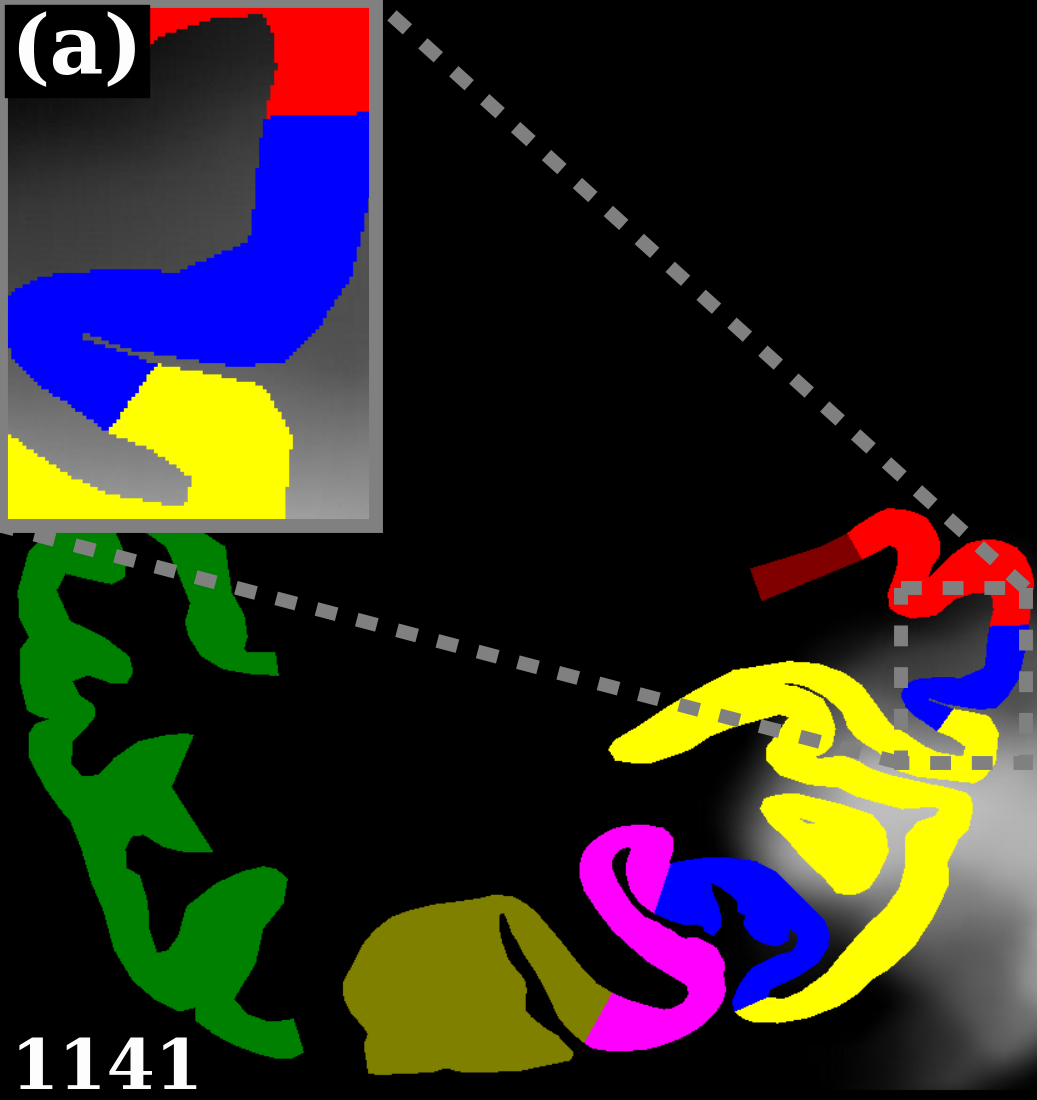}
	\end{subfigure}%
	\begin{subfigure}[t]{.19\linewidth}
		\centering
		\includegraphics[width=.97\linewidth]{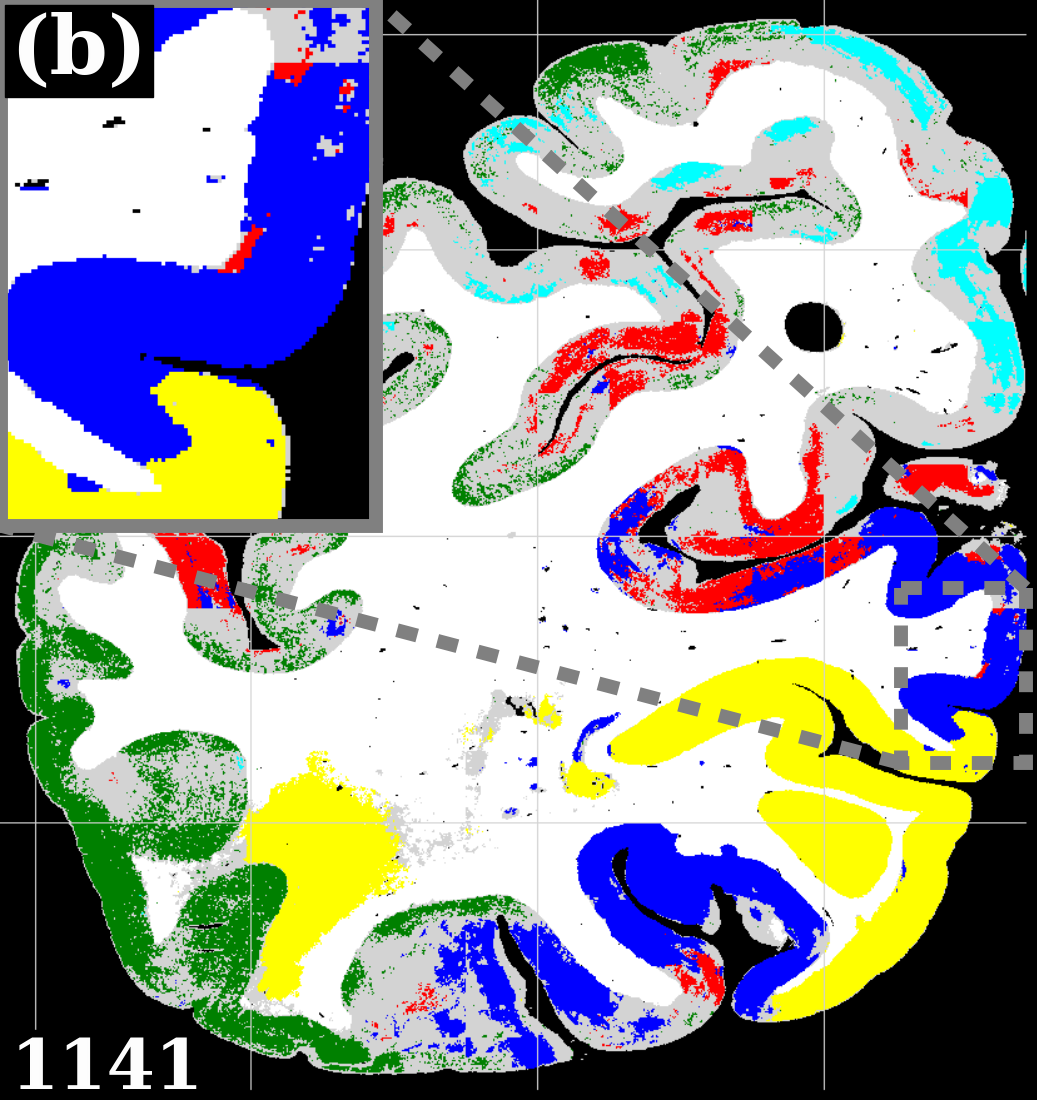}
		%split_alternate_B01_all/laplace_rotation/no_depth/gray_corrected_cellseg
	\end{subfigure}%
	\begin{subfigure}[t]{.19\linewidth}
		\centering
		\includegraphics[width=.97\linewidth]{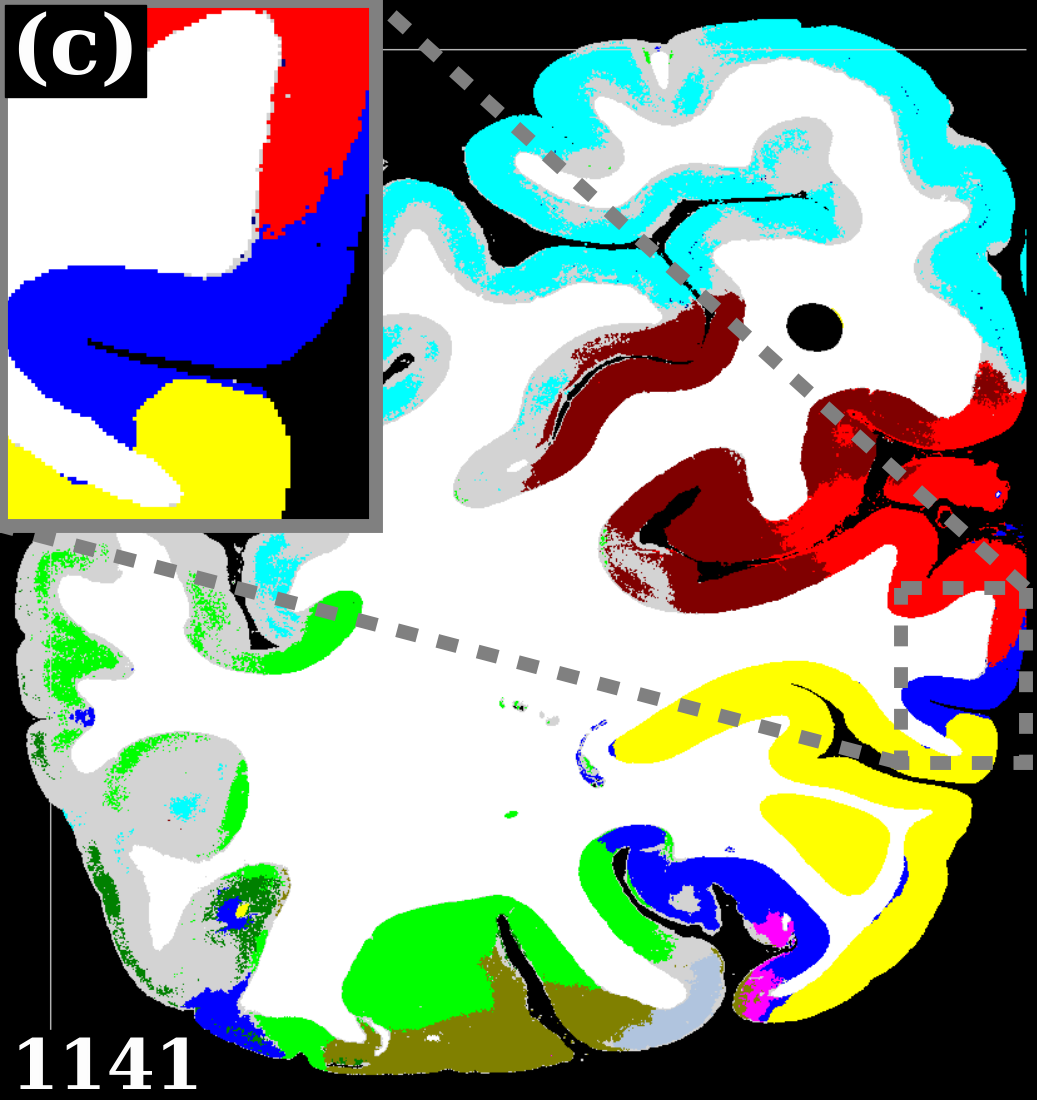}
		%split_alternate_B01_all/fusion_jubrain_finetuning/laplace_rotation/pmap
	\end{subfigure}%
	\begin{subfigure}[t]{.19\linewidth}
		\centering
		\includegraphics[width=.97\linewidth]{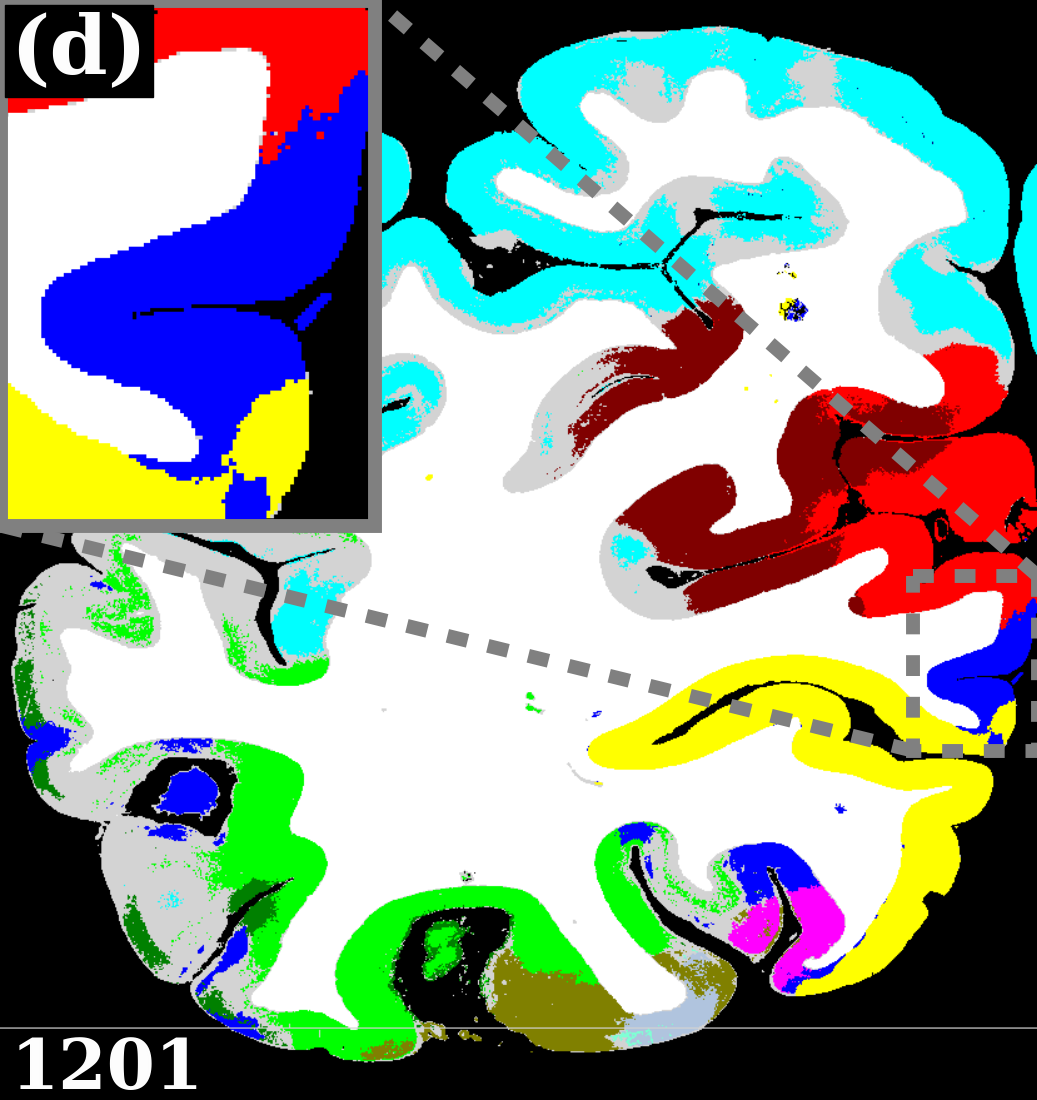}
		%split_alternate_B01_all/fusion_jubrain_finetuning/laplace_rotation/pmap
	\end{subfigure}%
	\begin{subfigure}[t]{.19\linewidth}
		\centering
		\includegraphics[width=.97\linewidth]{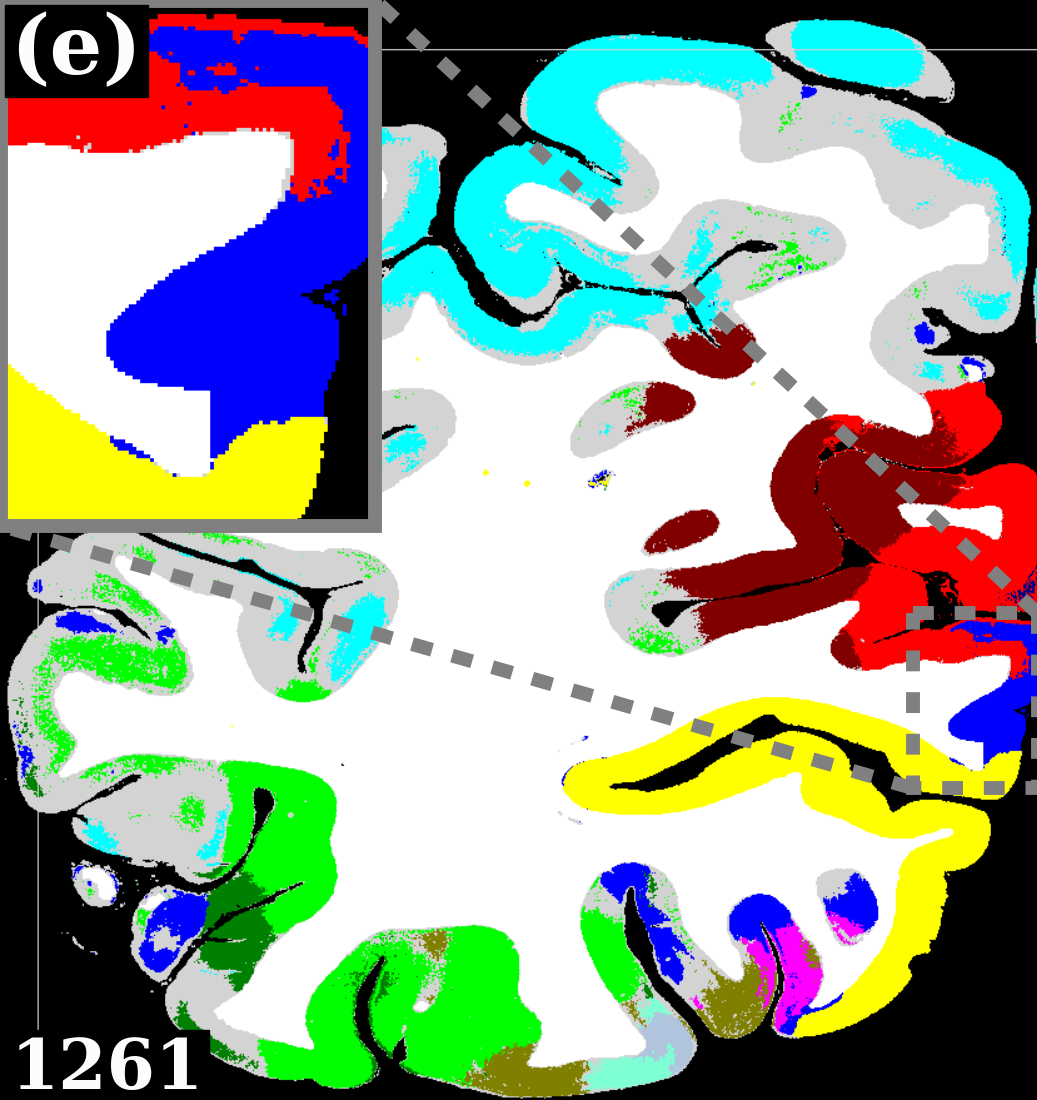}
		%split_alternate_B01_all/fusion_jubrain_finetuning/laplace_rotation/pmap
	\end{subfigure}
	\caption{
		\textbf{Qualitative evaluation of base and atlas-aware models},
		with (a) annotated areas (color) and projected probability of area hOc1 from the atlas (gray),
		(b) segmentation with the base model (no atlas prior), 
		and (c) - (e) segmentations with the atlas-aware model on consecutive sections. 
		Note that the confusion of non-labeled cortex with area FG4 is to be expected, because of all annotated areas, FG4 is closest to surrounding cortical areas. %and the correct cytoarchitectonic label has not been part of the classification.
		For the color coding see Fig.~\ref{fig:hist}.
	}
	\label{fig:res}
\end{figure*}

\begin{figure}[t!]
	\centering
	\begin{subfigure}{.32\columnwidth}
		\centering
		\includegraphics[width=.96\linewidth]{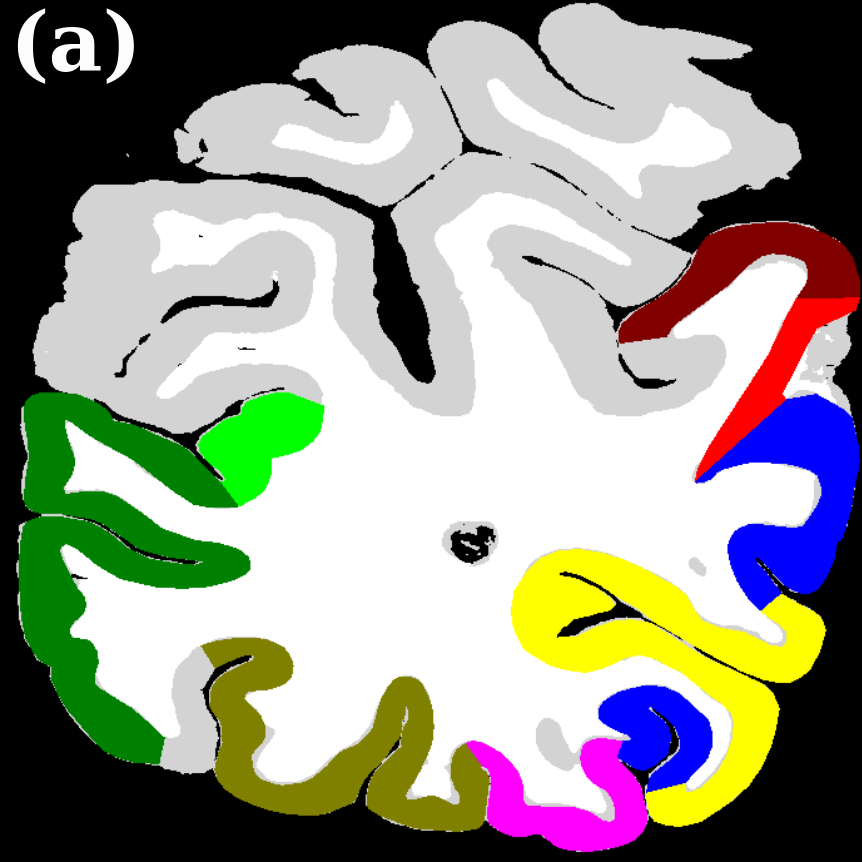}
	\end{subfigure}%
	\begin{subfigure}{.32\columnwidth}
		\centering
		\includegraphics[width=.96\linewidth]{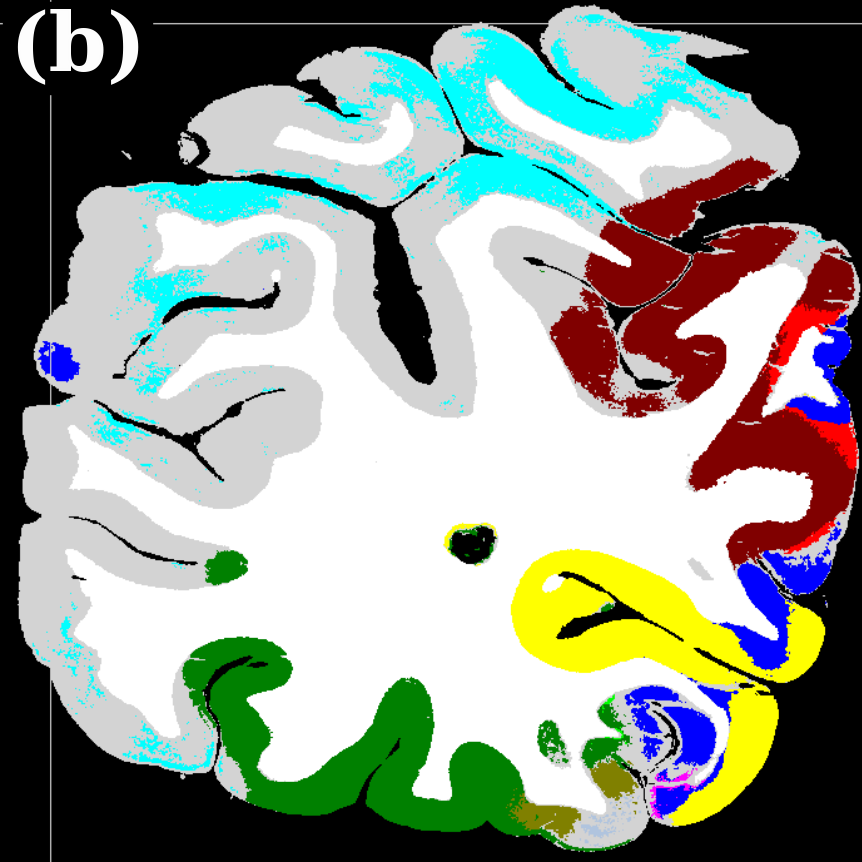}
		%split_alternate_B01_B02_B03_B10/fusion_jubrain_finetuning/random_rotation/pmap_gray
	\end{subfigure}%
	\begin{subfigure}{.32\columnwidth}
		\centering
		\includegraphics[width=.96\linewidth]{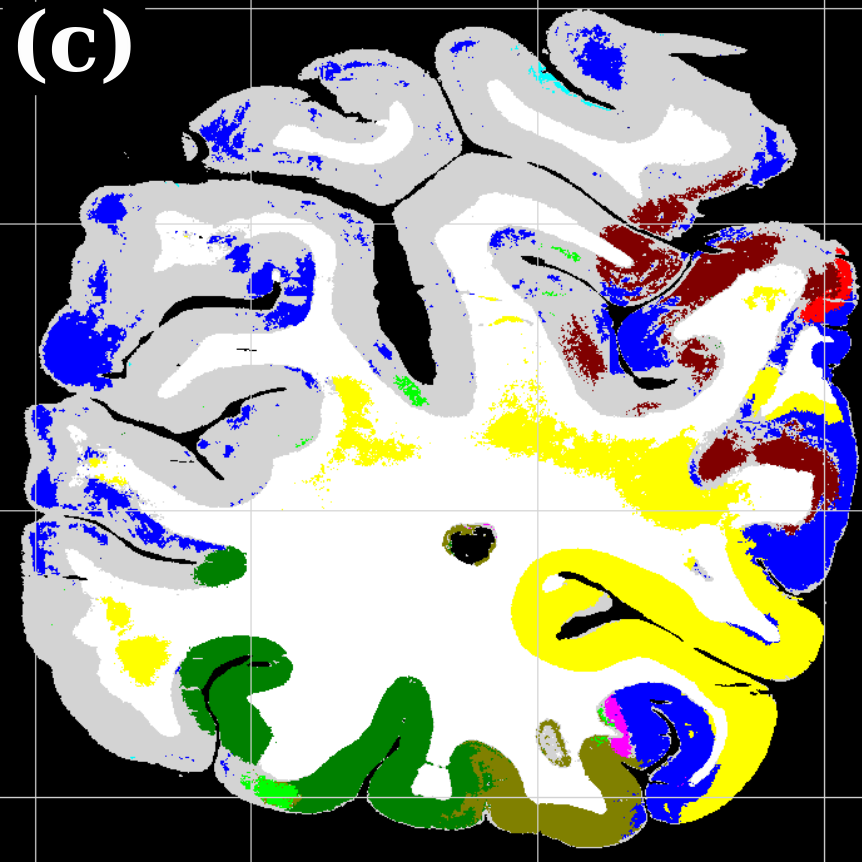}
		%split_alternate_B01_B03_B10/fusion_jubrain_finetuning/pmap_gray_depth
	\end{subfigure}%
	\caption{
		\textbf{Transferability to new brains}, 
		with (a) groundtruth,
		(b) predictions of a model trained on all brains ($\epsilon=10.0$),
		and (c) predictions of a model trained on three of the four brains in the dataset ($\epsilon=14.2$). 
		The displayed section is from the brain excluded in (c). 
		The error $\epsilon$ is computed for the displayed section.
		For the color coding see Fig.~\ref{fig:hist}.
	}
	\label{fig:transfer}
\end{figure}

\vspace*{-.5\baselineskip}
\subsection{Influence of atlas data and anatomical correctness}
\label{sec:res:correct}
\vspace*{-.5\baselineskip}

To evaluate the contribution of the global anatomical information (atlas prior, Sec.~\ref{sec:arch:arch}), we trained one model without atlas information (base) and compared it to our model including atlas information (atlas-aware). 
The base model predicts the most frequent areas (hOc1, hOc2, hOc4la) with good precision (Fig.~\ref{fig:res}b), but only the atlas-aware model manages to predict areas which are not as much represented in the dataset (Fig.~\ref{fig:res}c, and confusion matrices in Fig.~\ref{fig:cm}). 
%The confusion matrices (Fig.~\ref{fig:cm}) support this; the matrix of the atlas-aware model has a more populated diagonal. 
In general, the model performs better on more frequent areas, yielding higher $D_C$ scores. 
The error $\epsilon$ of the atlas-aware model drops $2$ points and the $D_C$ score rises by $0.1$ compared to the base model.
%, indicating that including the atlas prior helps the net to make more anatomically correct predictions. 
The results indicate that the network indeed learns to resolve labelling errors by exploiting the atlas prior to limit the number of possible labels per example.

\vspace*{-.5\baselineskip}
\subsection{Spatial consistency and transferability to other brains}
\label{sec:res:transfer}
\vspace*{-.5\baselineskip}

% --- spatial consistency ---
Figures~\ref{fig:res}c-d show predictions of the atlas-aware model on three sections with distance $1.2\,mm$. 
Although the model has no direct knowledge of spatial inter-slice dependencies, the predictions are consistent in the z-direction. 
In particular the boundary between hOc1 and hOc2 is consistent. 

% --- transferability ---
To see how well our model generalizes w.r.t.~different subjects, we trained a new model with one particular brain excluded from training, and evaluated on the latter (see Fig.~\ref{fig:transfer}).
%Fig. \ref{fig:transfer} shows the segmentation of one validation slice from the excluded brain made by the model excluding the brain (\ref{fig:transfer:noB02}) and by the model including the brain during training (\ref{fig:transfer:withB02}). 
This model predicts more frequent visual areas reasonably well, and the error $\epsilon$ only rises $4$ points to $14.2$ compared to a model trained on all brains. 
This suggests that our model is in principle transferable to previously unseen brains.

\section{Conclusion}
% --- summary ---
We presented a model that predicts visual areas on high-resolution histological sections exploiting both texture features and probabilistic atlas information.
The predictions are spatially consistent and reproducible on sections of previously unseen brains. 
We have shown that a probabilistic atlas prior has a positive effect on the model performance.
In a straightforward two-step process we generated accurate gray/white matter segmentations from a few training data points. 
% --- future work ---
In future work we plan to extend this model to include more cortical areas, and further study generalization across subjects.
Possible improvements include enforcing topological constraints, and exploiting 3D information as provided in reconstructed volumes.

\vspace*{\baselineskip}
\noindent{\textbf{Acknowledgements:}
\small
	This work was partially supported by the Helmholtz Association through the Helmholtz Portfolio Theme “Supercomputing and Modeling for the Human Brain”, and by the European Union’s Horizon 2020 Framework Research and Innovation under Grant Agreement No. 7202070 (Human Brain Project SGA1).
}
% To start a new column (but not a new page) and help balance the last-page
% column length use \vfill\pagebreak.
% -------------------------------------------------------------------------

% References should be produced using the bibtex program from suitable
% BiBTeX files (here: strings, refs, manuals). The IEEEbib.bst bibliography
% style file from IEEE produces unsorted bibliography list.
% -------------------------------------------------------------------------
\bibliographystyle{IEEEbib}

\bibliography{references}

\end{document}